\documentclass{article}

\usepackage[preprint]{corl_2021} 

\usepackage{amsmath}
\usepackage{amsfonts} 
\usepackage{graphicx}
\usepackage[ruled,vlined]{algorithm2e}
\usepackage{algorithmic}
\usepackage{subcaption}
\usepackage{enumitem}
\usepackage{changepage}
\usepackage{booktabs}
\usepackage{wrapfig}
\usepackage{xcolor}
\hypersetup{citecolor=blue}
\hypersetup{linkcolor=blue}
\hypersetup{colorlinks=true}

\title{Learning Vision-Guided Dynamic Locomotion Over Challenging Terrains}

\author{
  Zhaocheng Liu,~~Fernando Acero,~~Zhibin Li\\
  Institute of Perception, Action and Behaviour\\
  School of Informatics\\
  University of Edinburgh\\
  \texttt{\{zhaocheng.liu, fernando.acero, zhibin.li\}@ed.ac.uk} 
\vspace{-1.8em}
}

\begin{document}
\maketitle

\begin{abstract}
    Legged robots are becoming increasingly powerful and popular in recent years for their potential to bring the mobility of autonomous agents to the next level. This work presents a deep reinforcement learning approach that learns a robust Lidar-based perceptual locomotion policy in a partially observable environment using Proximal Policy Optimisation. Visual perception is critical to actively overcome challenging terrains, and to do so, we propose a novel learning strategy: Dynamic Reward Strategy (DRS), which serves as effective heuristics to learn a versatile gait using a neural network architecture without the need to access the history data. Moreover, in a modified version of the OpenAI gym environment, the proposed work is evaluated with scores over \textbf{90\%} success rate in all tested challenging terrains. 
\end{abstract}



\section{Introduction}
In recent years, legged robots have gained popularity in academic and industrial settings \cite{bostonBigDog}, due to their potential uses in collaboration with humans in complex environments, as well as substituting humans to perform tasks in hazardous environments. Existing approaches used in legged locomotion can be categorised into: \textit{optimization-based controllers}, which typically make use of mixed-integer nonlinear optimization \cite{MIT_Opti, RPC} and Model Predictive Control (MPC) \cite{MPC_Guiyang, NMPC, Angelini_MPC}; and \textit{learning-based controllers}, frequently trained using Reinforcement Learning (RL) \cite{DRL_Haarnoja, yang2018learning, DRL_Xie, ETH} or Imitation Learning (IL) \cite{ETH_Imitation, SonyImitation}. In a nutshell, the optimization-based controllers are usually more predictive but less resilient, whereas the learning-based controllers usually have a large training overhead.

The motivation for developing learning-based control policies for legged locomotion is multifaceted. It can be argued that learning-based approaches are not constrained by human design in the same way as optimization-based approaches, and they do not require the same level of tuning of control parameters by human experts \cite{HighMPC}. Moreover, learning-based controllers have a significant advantage over optimization-based approaches in that they require significantly less computation time on deployment, particularly in complex scenes and terrains where optimization-based approaches may not be able to solve for the control inputs online in real-time \cite{MIT}, as they capture the experience necessary to perform legged locomotion during the training process. 

Several contributions have demonstrated successful transfer of policies trained in simulation into reality, such as \cite{Google_MLP_SimToReal_minitaurs} which demonstrated the capability to train a locomotion policy using Deep RL to later deploy it on a real quadrupedal robot, and proposes techniques to achieve good simulation-to-reality transfer. Recent contributions on learning-based control for legged locomotion have demonstrated that such approaches can outperform optimization-based control in multiple settings. For example, a multi-expert architecture can dynamically synthesizes several neural networks instead of one DRL-based network \citet{yang2020}, yielding a unified framework that can perform multiple tasks and learns to generate adaptive transitions between trotting, turning, perturbation resistance and fall recovery modes. 

However, a common characteristic of contributions up to date is that the control policy is ``blind", i.e. it does not have access to exteroceptive data such as vision or scene depth information. This fundamentally limits the capability of the policy to locomote over complex terrains that require the use of exteroceptive feedback, e.g. scenes featuring large gaps or obstacles. 


{\bf Related work:} \citet{stereoVision2} (2013) and \citet{siravuru2017deep} (2017) proposed methods of using cameras to predict the safe footholds which includes a large CNN processing cost. \cite{song2018recurrent} used a recurrent policy to encode directly the visual perception from LiDAR sensing and proprioceptive information, and \citet{cnnLocomotion} proposed a Convolutional Neural Networks based classifier to plan optimal quadrupedal footholds. Recently, \citet{ox} (2021) proposed a hybrid method of RL and trajectory optimization algorithm that takes terrain height map to generate an adaptive joint-level trajectories, and a similar DRL approach was proposed by \cite{DRL_Zero_shot}. \citet{tsounis2020deepgait} (2020) proposed a ``DeepGait'' locomotion model based on the terrain heightmap given by the simulation environment. We introduce an on-board LiDAR sensor to capture the terrain information which is more practical and efficient to obtain in the real world.  

\label{section:DRL} {\bf Background:} Deep RL algorithms are based on the \textit{Markov Decision Process (MDP)} paradigm \cite{PPO, DRL, MDP}, which formalizes the process by which an agent learns to map state observations into actions. The MDP problem is defined as $\{S,A,\mathcal{T},\mathcal{R},\gamma,s_0\}$, where S and A are sets of continuous states and actions, and $s_0$ is the initial state. The transition function $\mathcal{T}: S \times A \times S'\rightarrow \mathbb{R}$ represents the transition probabilities between states by taking an given action, and the reward function $\mathcal{R}: S \times A \rightarrow \mathbb{R}$ provides the immediate reward upon taking an action at a state. The objective can be stated as learning a policy $\pi_{\theta}: S \rightarrow A$ parametrized by $\theta$, that maximizes $\sum^{H}_{t=0} \gamma^{\,t} r_{t}$ where $\gamma$ is a discount factor, over a fixed horizon $H$. 

In order to train the policy, \textit{Proximal Policy Optimization} (PPO) \cite{PPO} is used. PPO is an on-policy RL algorithm that can be used for discrete and continuous control tasks, defining a policy update that makes use of first-order methods. PPO intends to update the policy weights with the biggest improvement step without accidentally causing collapse in performance of the policy, for which a clipped objective function is employed. It is for these reasons and its simplicity that PPO was chosen to train our locomotion policy.

{\bf Approach:} It is in this setting that we frame the motivation for our work, which aims to serve as an initial concept to demonstrate how visually-aware locomotion polices might be developed for expanding the current capabilities of legged robots in complex scenes. Specifically, the contributions of this paper are threefold: (i) the \textit{OpenAI Gym BipedalWalker} environment is modified to feature a quadrupedal robot with joint position control and access to exteroceptive depth data, and a baseline control policy is trained in this environment and presented, (ii) a conditional reward function named Dynamic Reward Strategy (DRS) is presented and identified as a key factor contributing to efficient and smooth learning of a galloping gait at a steady velocity, and (iii) the introduction of vision data into the policy observations results in a policy that can \textit{actively} overcome large gaps and obstacles, anticipating their presence and adapting the gait accordingly, in contrast to previous work in legged locomotion.


            

    \begin{figure}[h]
        \centering
        \includegraphics[width=0.85\textwidth]{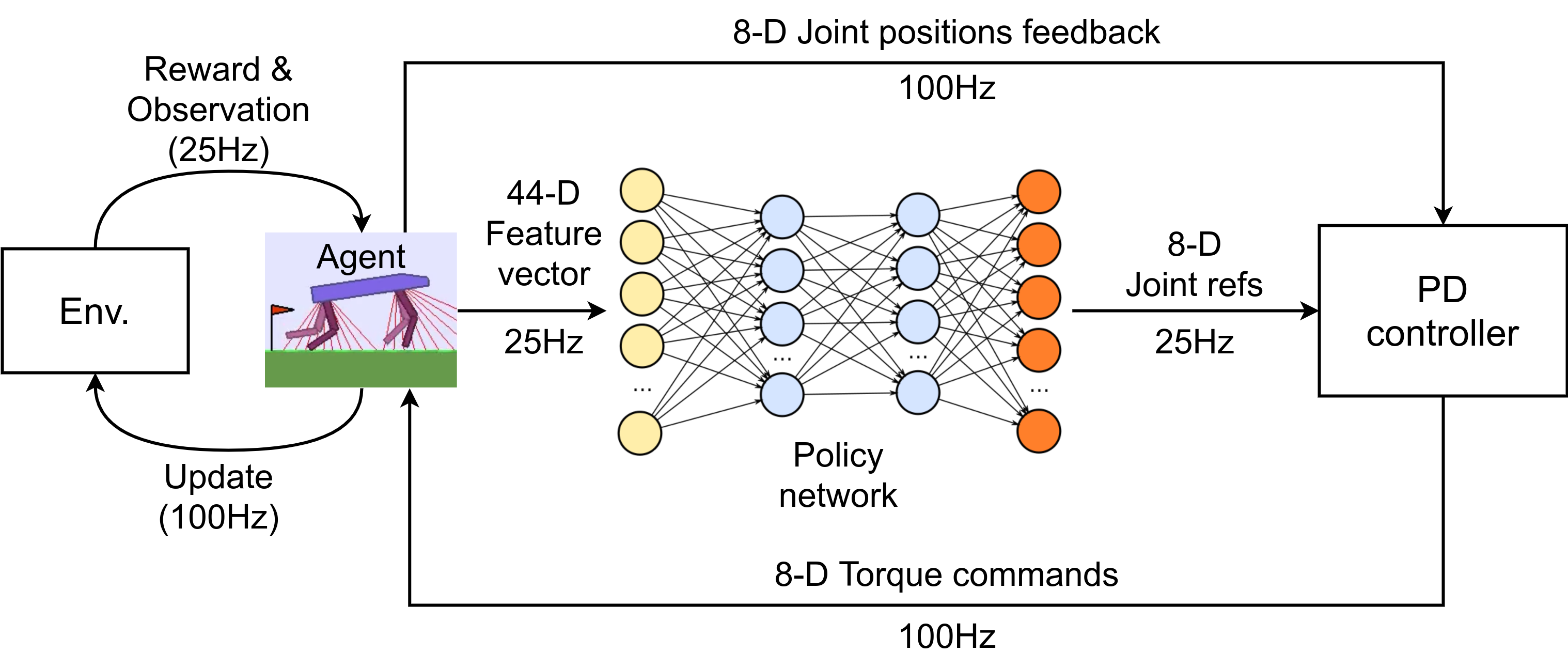}
        \caption{The rollout pipeline. The agent receives a 44-D observation vector from the environment, which is fed into to the policy network for the next desired actions. Then, the closed-loop impedance control takes these outputs of the policy network and applies moderate PD gains to actuate motors for acting in a compliant manner with the environment \cite{li2016compliance}. The simulator updates the environment and provides new observations to the agent, and the whole process is repeated and iterative simulated.}
        \label{fig:pipleline}
    \end{figure}
    
\clearpage
\section{Training Methodology}
    The training starts with a randomly initialised policy network $\pi_{\theta}$. During the training process, the PPO algorithm collects the trajectory data from the simulation, as shown in the Figure \ref{fig:pipleline}, and updates the policy with the training data. At each timestep, the agent receives an observation consisting of: (1) a 24-D proprietary state vector and (2) a 20-D LiDAR measurement vector as vision, where the proprietary state consists of: (a) a 4-D linear\slash angular displacement\slash velocity of the body, (b) an 16-D joint configuration\slash velocity vector and (c) a 4-D ground contact bitmap. Taking such observation, the policy learns the joint reference commands at a control frequency of 25 Hz, whereas the corresponding torques are generated by a low-level impedance control using moderate PD gains running at 100 Hz. The hyperparameters used in training can be found in Appendix \ref{apx:hyperparm}.

    
            
    

\section{Dynamic Reward Strategy}
    This section introduces the design of the reward function, and the \textit{Dynamic Reward Strategy} (DRS) which is the key factor for learning smooth quadrupedal locomotion in this environment. We propose a reward function:
    \begin{align}
        \text{reward}_{t} &= \begin{cases}
            \mathbf{\mathrm{P}}_{fall},& \text{if has felt down} \\
            c_1 (c_2 - c_3 (v_x - v_x^{*})^{2}) + \|\mathbf{k}_1 \circ \tau\|_{1} +  \|\mathbf{k}_2 \circ \ddot{\mathbf{q}}\|^{2}_{2}, & \text{otherwise}
        \end{cases} 
        \vspace{1em} \\
        \mathbf{k}_1 &= k_{1,\;i \in [1\;\ldots\;N]} = \begin{cases}
            c_4,& \text{if the related foot is in contact, for leg $i$} \\
            c_5,& \text{otherwise}
        \end{cases} \\ 
        \mathbf{k}_2 &= k_{2,\;j \in [1\;\ldots\;N]} = \begin{cases}
            c_6,& \text{if the related foot is in contact, for leg $j$} \\
            c_7,& \text{otherwise}
        \end{cases}
    \end{align}
    
    Where, $c_1, \;\ldots\;, c_7$ and $\mathbf{\mathrm{P}}_{fall}$ are constants, $\tau$ is the torque command and $\dot{\mathbf{q}}$ is the joint angular acceleration. The falling penalty $\mathbf{\mathrm{P}}_{fall} < 0$ is applied whenever the agent's body touches the terrain. The symbol ``$\circ$'' denotes the element-wise vector multiplication. $\|\mathbf{k}_1 \circ \tau\|_{1}$ is the joint torque penalty and $\|\mathbf{k}_2 \circ \ddot{\mathbf{q}}\|^{2}_{2}$ is the joint angular velocity penalty, $\mathbf{k}_1$ and $\mathbf{k}_2$ are dynamic penalty coefficient vectors. 
    
    {\bf Desired velocity.} The term $c_1 (c_2 - c_3 (v_x - v_x^{*})^{2})$ encourages the agent to have a steady horizontal velocity at $v_x^{*}$~m/s, the reward decreases when the velocity deviates from the desired velocity and becomes an increasing penalty.
    
    {\bf Joint angular acceleration penalty.} The stochastically learned policies often produce high frequency oscillations at the joints, which is is characterised by joint velocities $\Delta \dot{\mathbf{q}}$ that frequently flip sign. Formally, we have
    \begin{equation}
        \Delta \dot{\mathbf{q}} \propto \int_{t}^{t+\delta t} \ddot{\mathbf{q}}\;dt
    \end{equation}
    In order to avoid high frequency oscillations, a quadratic angular acceleration $\text{penalty}_{t} \propto \|\ddot{\mathbf{q}}\|^{2}_{2}$ is equally applied to every joint. This removes the shaky joint trajectory but also penalizes the swing phase of locomotion. Because of this, a more sophisticated reward function is needed.
    
    {\bf Dynamic Reward Strategy (DRS).} On top of a well-defined reward function, in order to learn a better locomotion gait, we propose the DRS (Algorithm \ref{algo:DRS}). DRS promotes a high power output when the leg is in contact, and reduces the high frequency motions and wastage of energy in the legs during the leg swing phases. Further speaking, this strategy guides the policy to learn the following properties: (i) smooth swings of the leg for better foot placements, (ii) fast and smooth leg swings in the air, with small angular acceleration, (iii) a greater ground reaction force to reduce slip and (iv), a burst of ground reaction force for better jumping capability. 
    \label{section:desired_gait_property}

    The joint torques are always penalised to minimize power efficiency. Since the joint angular acceleration is directly proportional to the joint torque, torque penalty coefficients are proportional to the joint angular acceleration penalty coefficients.
    
    \begin{algorithm}[h]
        \SetAlgoLined
        \caption{Dynamic Reward Strategy} 
        \label{algo:DRS}
        \DontPrintSemicolon
        
        \For{each leg, $i$, in the quadrupedal robot}{
            \uIf{the lower leg is in contact to the terrain}{
                Decrease the joint torque and angular acceleration penalty coefficients $k_{1, i}$ and $k_{2, i}$ \;
            }\Else{\tcc{the lower leg is in swing phase}
                Increase the joint torque and angular acceleration penalty coefficients $k_{1, i}$ and $k_{2, i}$ \;
            }
        }
    \end{algorithm}
    
\section{Simulation Environment}

\subsection{Agent design}
        
    
    
\begin{figure}[h]
        \centering
        \begin{subfigure}[b]{.35\textwidth}
            \centering
            \includegraphics[width=\textwidth]{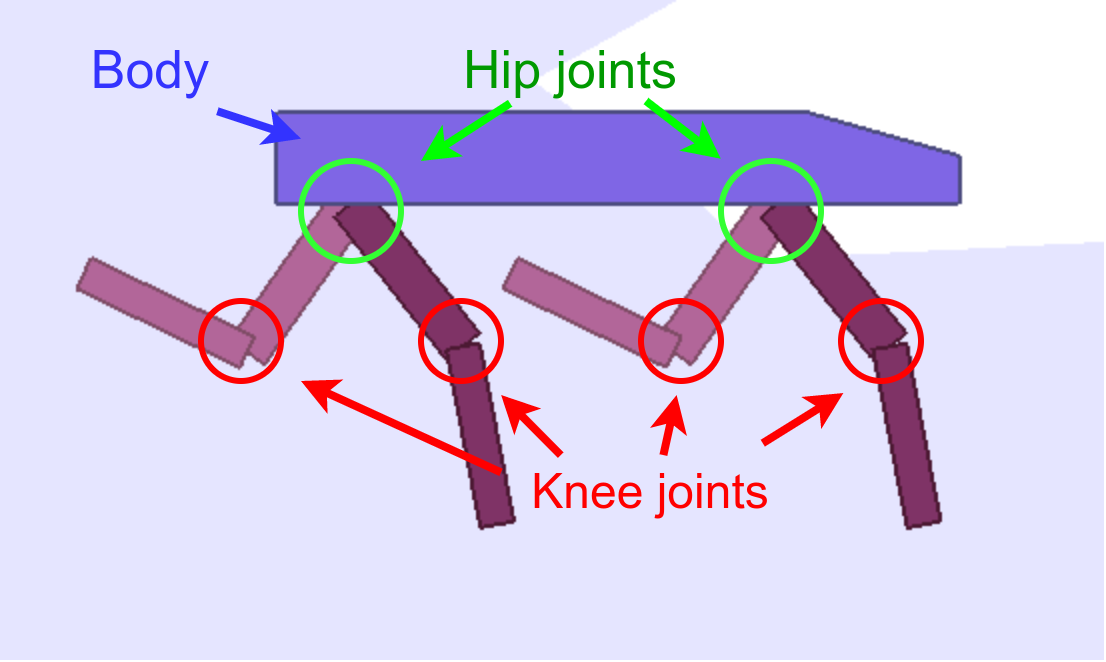}
            \caption{The agent}
            \label{fig:the agent}
        \end{subfigure}
        \hspace{2cm}
        \begin{subfigure}[b]{.35\textwidth}
            \centering
            \includegraphics[width=\textwidth]{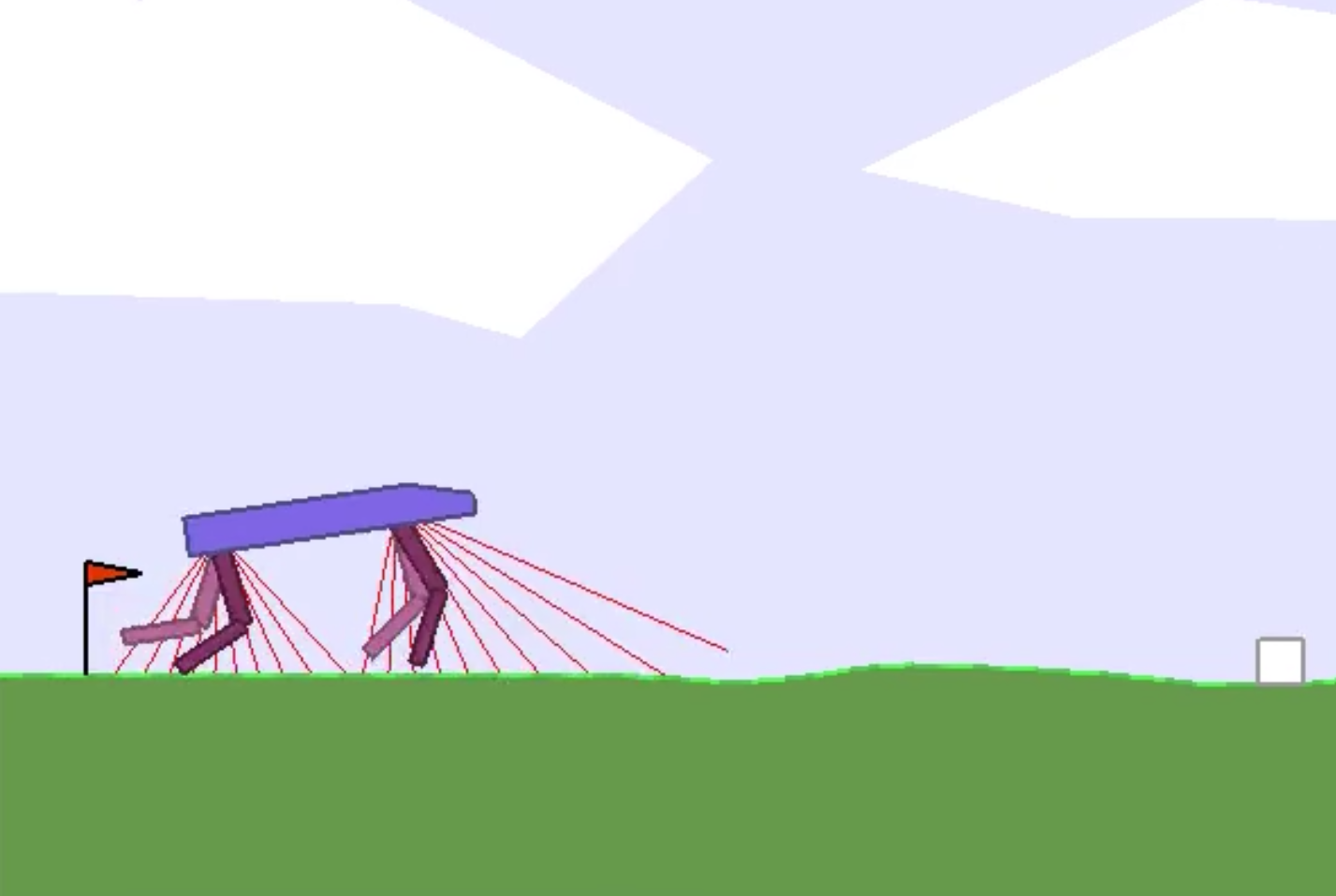}
            \caption{The agent in the environment}
            \label{fig:agent in env}
        \end{subfigure}
        
        \vspace{0.3cm}
        \begin{subfigure}[b]{0.4\textwidth}
            \centering 
            \includegraphics[width=\textwidth]{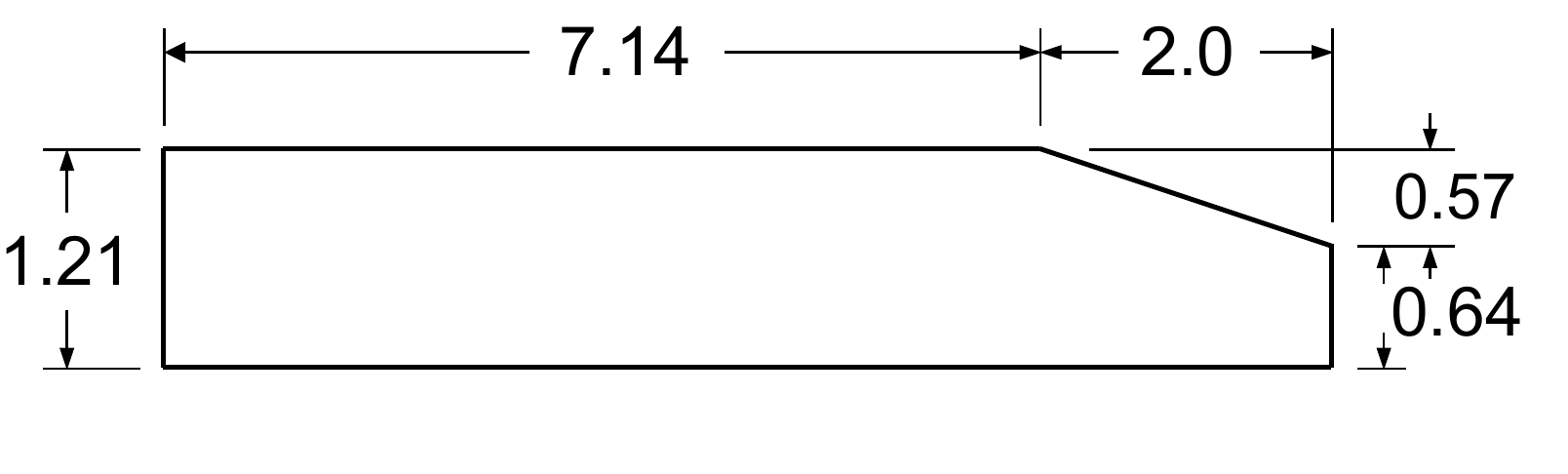}
            \caption{The body}
            \label{fig:the agent body}
        \end{subfigure}
        \hspace{1cm}
        \begin{subfigure}[b]{0.2\textwidth}
            \centering
            \includegraphics[width=\textwidth]{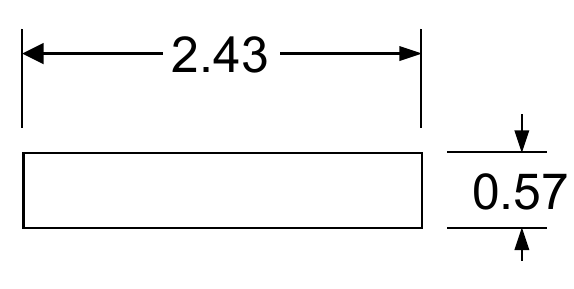}
            \caption{The upper leg}
            \label{fig:robot legs}
        \end{subfigure}
        \hspace{1cm}
        \begin{subfigure}[b]{0.2\textwidth}
            \centering
            \includegraphics[width=\textwidth]{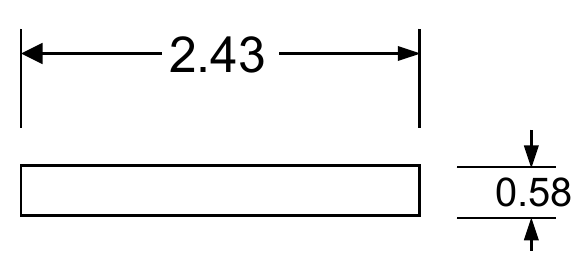}
            \caption{The lower leg}
            \label{fig:robot legs2}
        \end{subfigure}
        
        \begin{subfigure}[b]{0.3\textwidth}
            \centering \rule{0pt}{0.3cm}
            \includegraphics[width=\textwidth]{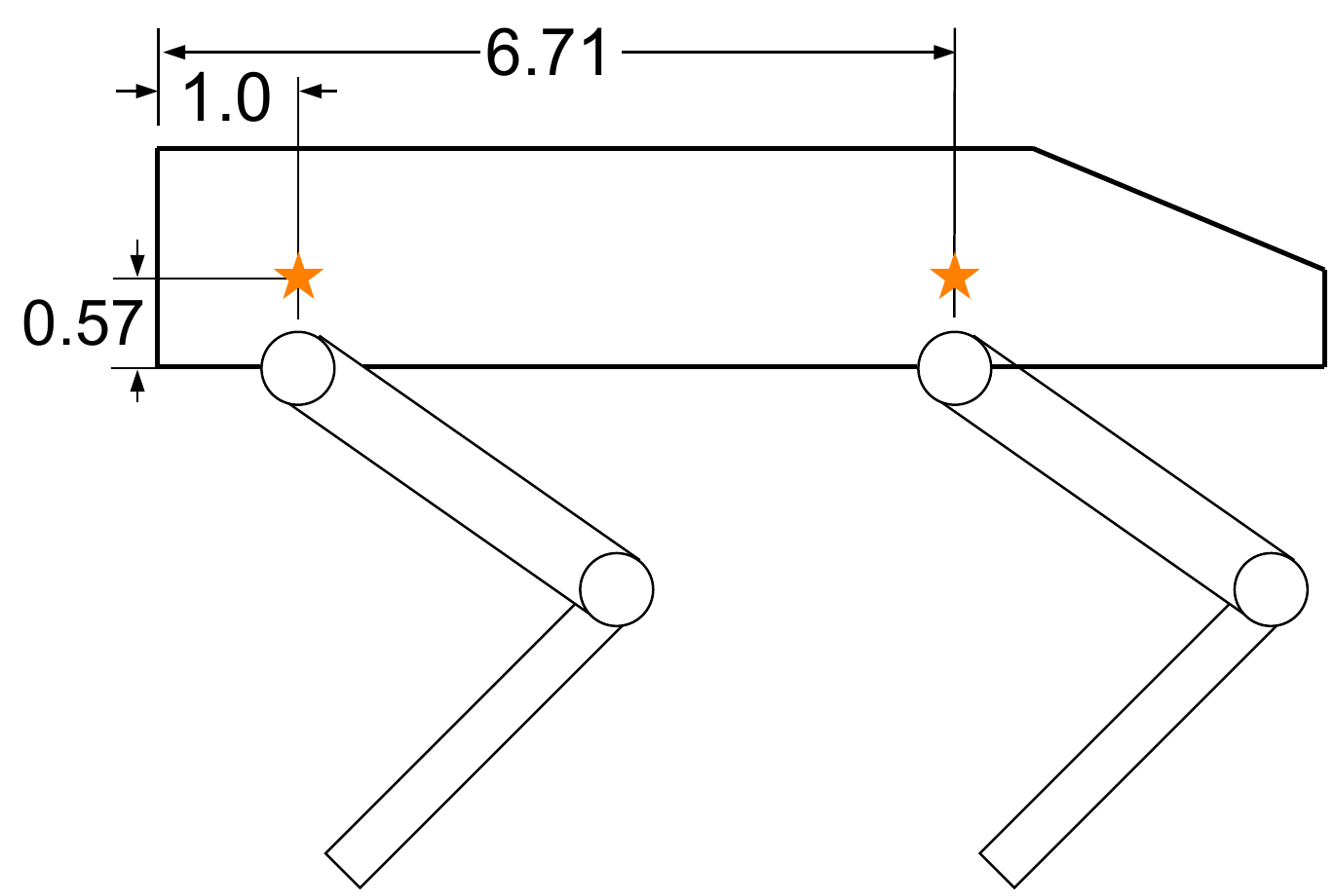}
            \caption{Design of the agent}
            \label{fig:robot combo1}
        \end{subfigure}
        \hspace{1cm}
        \begin{subfigure}[b]{0.25\textwidth}
            \centering
            \includegraphics[width=0.5\textwidth]{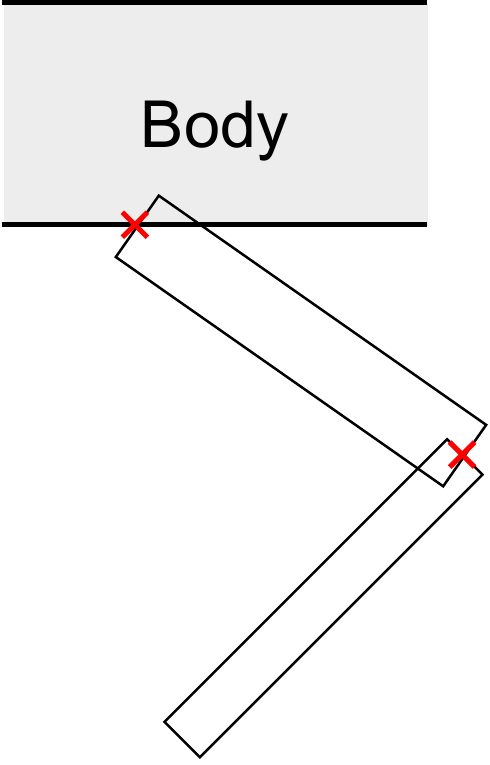}
            \caption{The connection points}
            \label{fig:robot combo2}
        \end{subfigure}
        \hspace{1cm}
        \begin{subfigure}[b]{0.25\textwidth}
            \centering
            \includegraphics[width=1.1\textwidth]{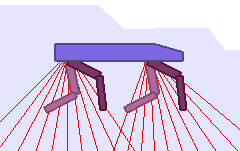}
            \caption{LiDAR configuration}
            \label{fig:lidar_config}
        \end{subfigure}
        
        \caption{The blueprint of the agent, in meters (m)}
        \label{fig:robot blueprint}
    \end{figure}

    The robot is modified based on the \textit{OpenAI BipedalWalker} model which has a unrealistic large size (in meters) but a reasonable weight (in kilograms). See in the Figure \ref{fig:the agent} and the blue print in the Figure \ref{fig:robot blueprint}. This is due to the embedded physics engine is lack of precision, especially when it involves complex collisions. In overview, the robot in this paper has 8~DoF: 4 hip joints and 4 knee joints body length of 9.1 m with leg lengths of 4.9 m and total weight of 13.6 kg. The LiDAR configuration is shown in the Figure \ref{fig:lidar_config}.
    
    \begin{figure}[h]
        \centering
        \subcaptionbox{}[.3\linewidth]
            {\includegraphics[height=3cm]{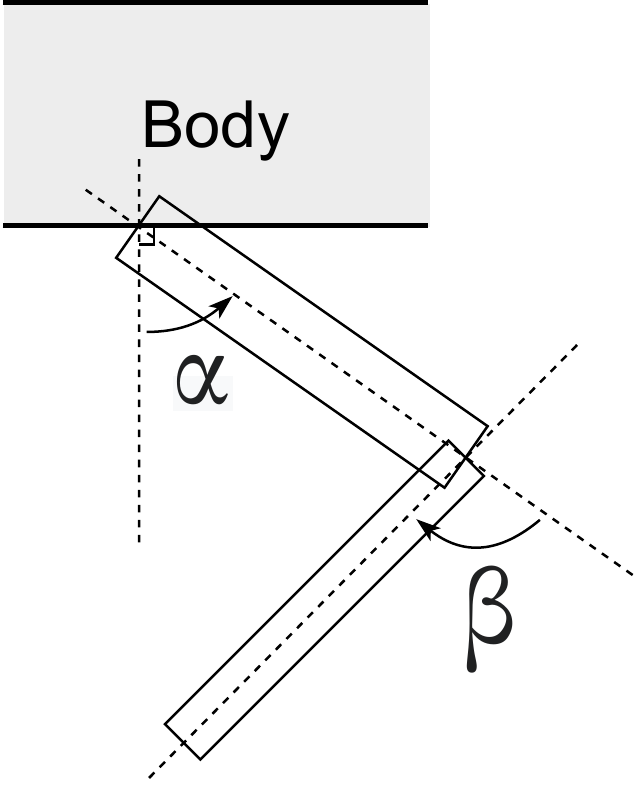}}
        \subcaptionbox{}[.3\linewidth]
            {\includegraphics[height=2.7cm]{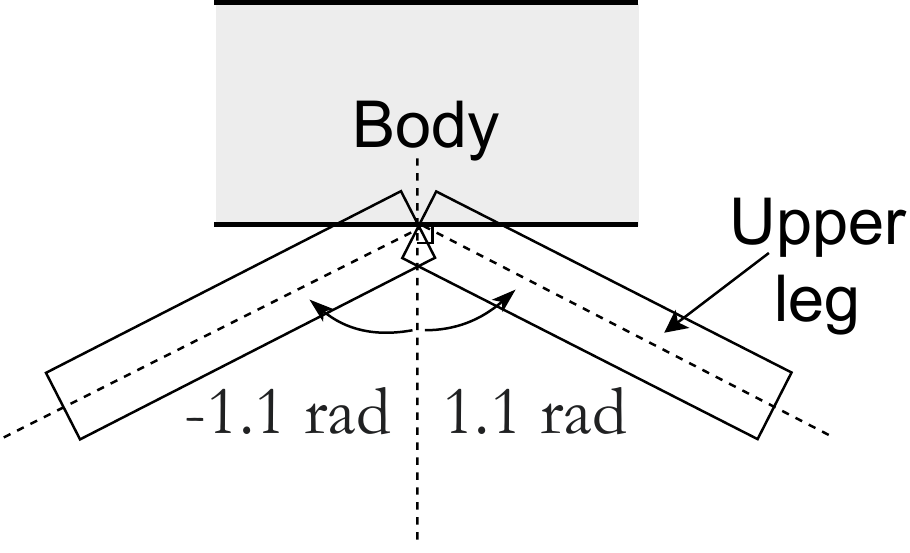}}
        \subcaptionbox{}[.3\linewidth]
            {\includegraphics[height=3.2cm]{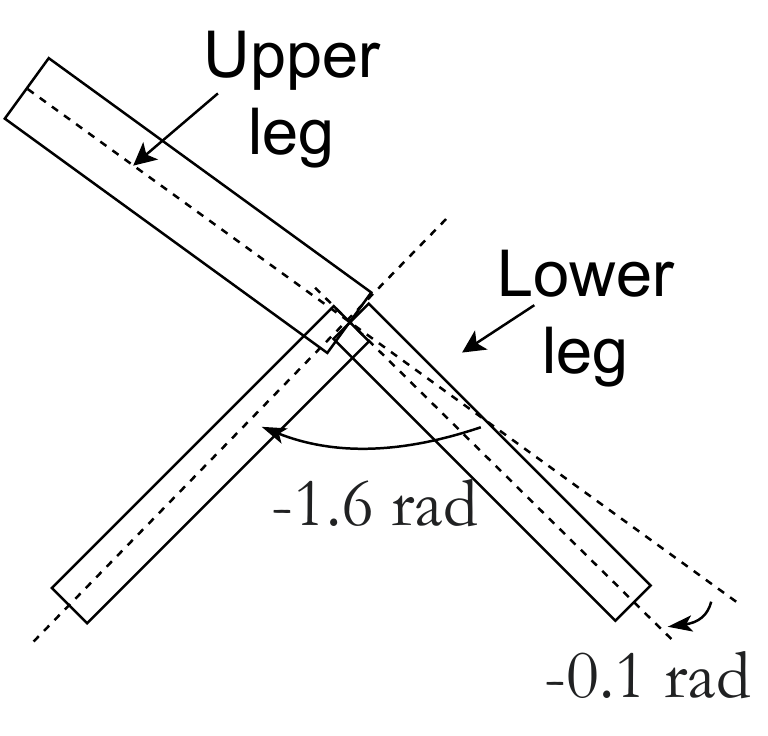}}
      
        \caption{Agent joint limits}
        \label{fig:jointAnglesAndLimits}
    \end{figure}
    
    The friction coefficient is set to 2.5 between the legs and the obstacles$\slash$terrain. This friction coefficient allows the agent to climb over the large stumps. On the other hand, it is still possible for the agent to make a slip. The mass of each links are: (i) The body is 11.47~kg. (ii) The upper leg is 0.30~kg, (iii) The lower leg is 0.24~kg, (iv) The total mass of the robot = $m_{body} + 4\times (m_{upper\; leg} + m_{lower\; leg}) $= 13.63~kg. The hip and knee joint positions are defined as $\alpha$ and $\beta$ (Figure \ref{fig:jointAnglesAndLimits}a) for each leg. And the joint limits are $\alpha \in [-1.1,1.1]$~rad (Figure \ref{fig:jointAnglesAndLimits}b) and $\beta \in [-1.6,-0.1]$~rad (figure \ref{fig:jointAnglesAndLimits}c). 
    
\subsection{Environment design}
    \begin{figure}[h]
        \centering 
        \begin{subfigure}[h]{1.\textwidth}
            \centering   \rule{0pt}{0.1cm}
            \includegraphics[width=\textwidth]{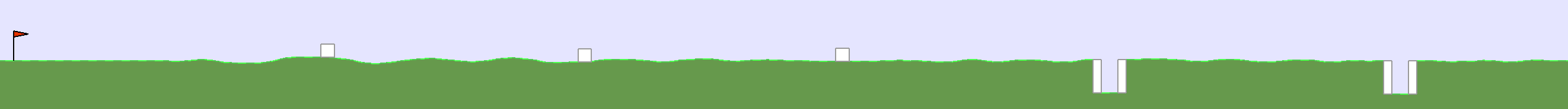}
            \caption{A rough terrain with obstacles \vspace{1em}}
            \label{fig:terrain hard}
        \end{subfigure}
        
        \begin{subfigure}[h]{.24\textwidth}
            \centering   \rule{0pt}{1.3cm}
            \includegraphics[height=2cm]{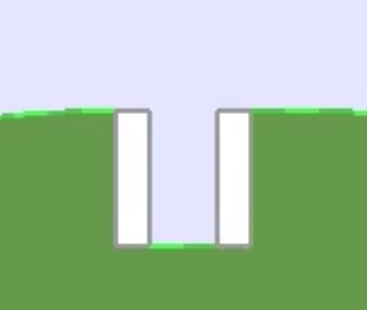}
            \caption{A gap}
            \label{fig:A gap}
        \end{subfigure}
        \begin{subfigure}[h]{.24\textwidth}
            \centering 
            \includegraphics[height=2cm]{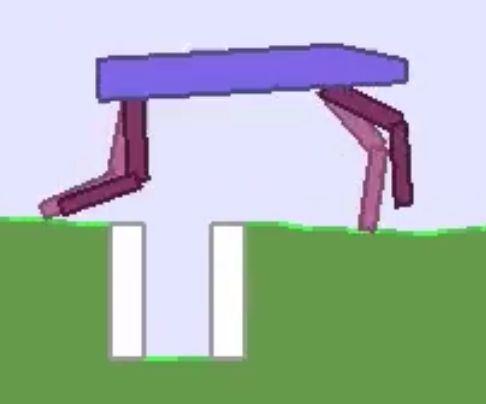}
            \caption{Agent at a gap}
            \label{fig:robot at gap}
        \end{subfigure}
        \begin{subfigure}[h]{.24\textwidth}
            \centering 
            \includegraphics[height=2cm]{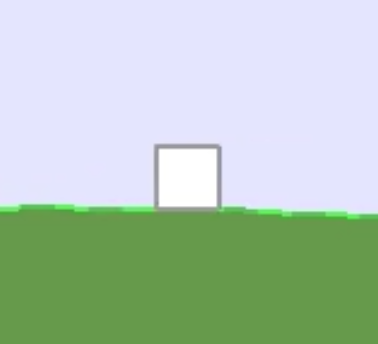}
            \caption{A stump}
            \label{fig:A stump}
        \end{subfigure}
        \begin{subfigure}[h]{.24\textwidth}
            \centering 
            \includegraphics[height=2cm]{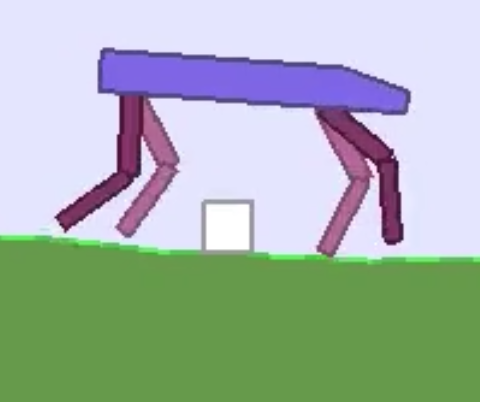}
            \caption{Agent at a stump}
            \label{fig:robot at stump}
        \end{subfigure}
        
        \label{fig:terrain combo}
        \caption{Representative examples of the rugged terrain and obstacles.}
    \end{figure}
    
    \label{sec:terrain}
    The experiments are conducted in a simulated environment, which is modified based on the \textit{OpenAI Gym BipedalWalker environment in 2-D}. The total length is about 93~m with 5 evenly spaced obstacles in the terrain, the obstacles are randomized in sizes, i.e. the length of the stump and the gap width. Figure \ref{fig:terrain hard} shows a sample of the rough terrain with obstacles. Obstacles are either a gap (Figure \ref{fig:A gap}) or a stump (Figure \ref{fig:A stump}), the proportion of body-to-obstacle size can be seen Figures \ref{fig:robot at gap} and \ref{fig:robot at stump} . Robots are initialized at the leftmost part of the terrain with a randomized small horizontal force applied to the body. 

\subsection{Forward Dynamics} 
    In this work, it is desired for the output of the control policy to be joint position reference commands, instead of joint torque commands, in order to simplify the mapping that the neural network has to learn between observations and actions. Due to the lack of support for Forward Dynamics (FD) computation in the original \textit{BipedalWalker} environment. As shown in the Algorithm  \ref{alg:getJointVel}, a FD algorithm is defined and integrated to the simulator, as the original environment only provides the dynamics for handling collisions between rigid bodies. 
    
    \begin{algorithm}[h]
        \DontPrintSemicolon
        
        \SetKwFunction{FMain}{GetJointVelocity}
        \SetKwProg{Fn}{Function}{:}{}
        \Fn{\FMain{$\dot{\theta}_{curr}$, $\mathcal{I}$, $\tau$}}{
            $\ddot{\theta} \; \leftarrow \; \tau \, / \, \mathcal{I}$ \;
            $\dot{\theta}_{new} \leftarrow$ clip($\dot{\theta}_{lower\;limit}$, $\dot{\theta}_{upper\;limit}$, $\ddot{\theta} \times \Delta t + \dot{\theta}_{curr})$ \;
            \KwRet $\dot{\theta}_{new}$ \;
        }
        \caption{Forward Dynamics algorithm}
        \label{alg:getJointVel}
    \end{algorithm}
    
    The FD algorithm implements the equation of $\dot{\theta}_{t+\Delta t} \approx \dot{\theta}_{t} + \ddot{\theta} \cdot \Delta t$ to calculate the joint velocity $\dot{\theta}_{new}$ of the next timestep, which is then clipped by the joint velocity limits: $[-4, 4] \;\text{rad}\cdot\text{s}^{-1}$ for the hip joints, and $[-6, 6] \;\text{rad}\cdot\text{s}^{-1}$ for the knee joints. 

\section{Results}
\label{section:novel eval}

\begin{figure}[h]
        \centering
            \begin{subfigure}[T]{.45\textwidth}
            \centering
            \includegraphics[width=\linewidth]{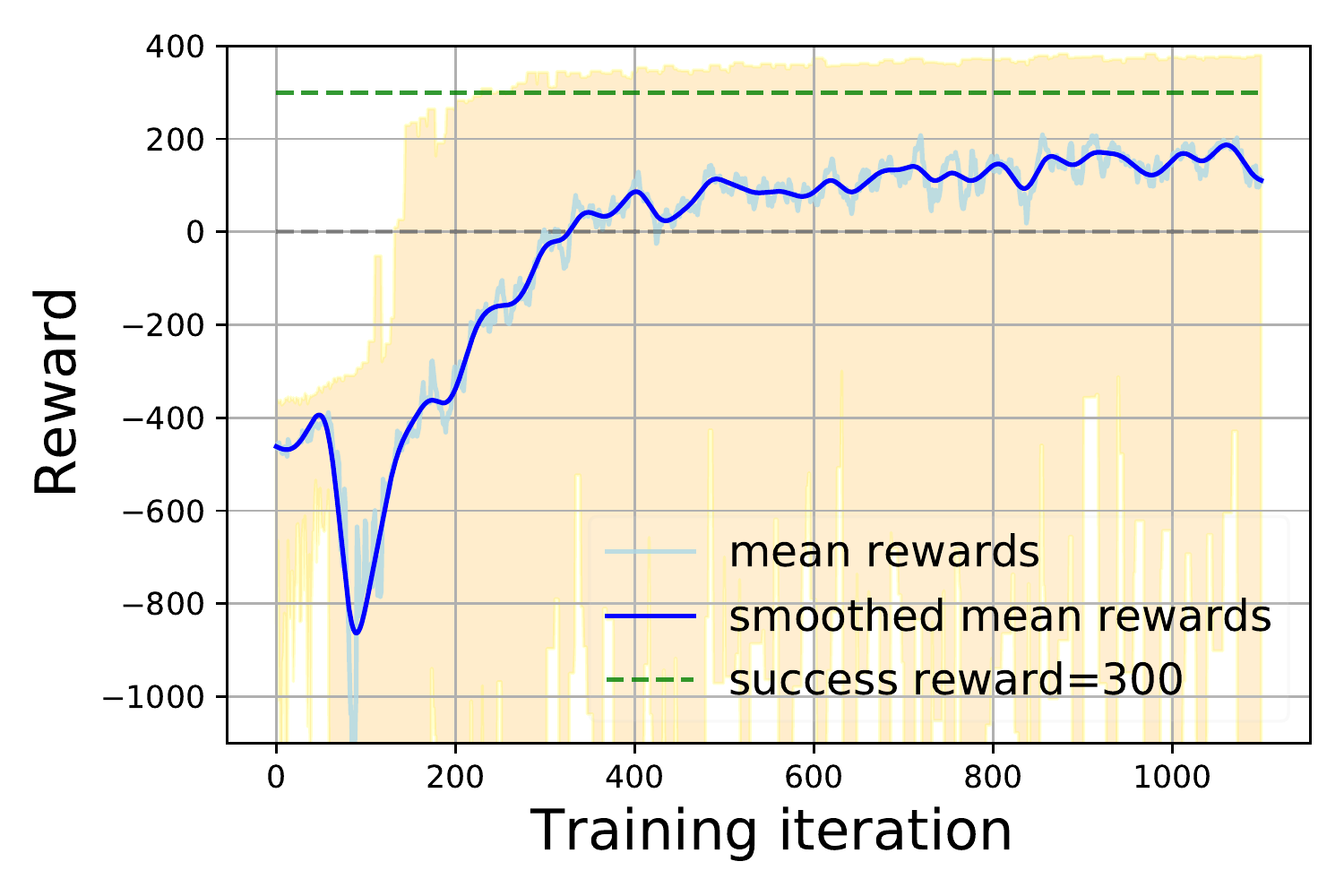}
            \caption{The training mean reward}
            \label{fig:novel results reward}
        \end{subfigure}
        \hspace{0.4cm}
        \begin{subfigure}[T]{.45\textwidth}
            \centering
            \includegraphics[width=\linewidth]{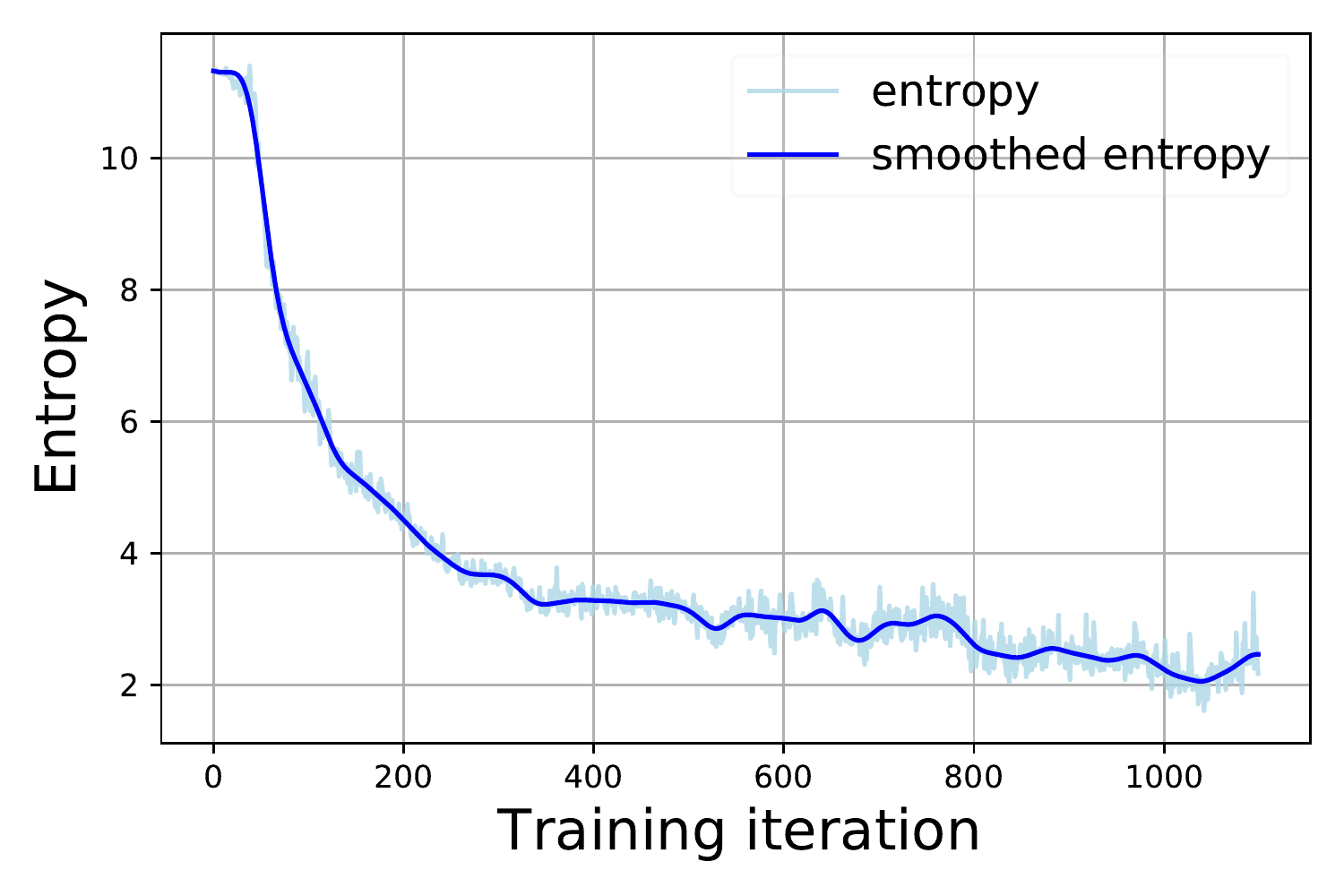}
            \caption{The training entropy}
            \label{fig:novel results entropy}
        \end{subfigure}
        
        \caption{Training results with DRS. {\bf The training reward curve} (Figure \ref{fig:novel results reward}), where the deep blue curve is the smoothed mean reward, the light blue curve is the raw mean reward, the green dotted line the success reward threshold, and the yellow shaded area is the range of the obtained rewards. Note that the policy samples actions stochastically in training mode in order to encourage exploration, but is sampled deterministically during testing. {\bf The policy entropy} (Figure \ref{fig:novel results entropy}) indicates the divergence of the policy during training. We consider that the policy has fully converged after about 800 iterations.}
        \label{fig:training result}
    \end{figure}

    With proposed method, the training was successful. The training result of the best experiment (Exp. 5) is shown in the Figure \ref{fig:training result} and the test results are shown in the Table \ref{tab:final}.
    
    \begin{table}[h] 
        \begin{adjustwidth}{-0.6cm}{}
        \centering
        \begin{tabular}{ccccccccc}
        \hline\hline 
        \textbf{Exp.} & \textbf{Target Vel.} & \textbf{Reward func.} & \textbf{Gait} & \multicolumn{2}{c}{\textbf{Rough terrain}} & \multicolumn{2}{c}{\textbf{Flat terrain}} \\
         & m/s & & & succ. rate & falling rate & succ. rate & falling rate \\
        \hline
        baseline & $[2, 0]^{T}$ & baseline & N/A & 0\% & 0\%  & N/A & N/A \\
        1 & $[2, 0]^{T}$ & static & crawling & 70\% & 2\%  & 65\% & 0\% \\
        2 &$[5, 0]^{T}$ & static & galloping & 0\% & 19\% & 0\% & 17\% \\
        3 & $[4, 0]^{T}$ & static & skipping & 85\% & 14\% & 94\% & 5\% \\
        4 & $v^{*}_{x} = 4$ & static & galloping & 72\% & 25\% & 83\% & 17\% \\
        5 & $v^{*}_{x} = 4$ & DRS & galloping & \textcolor{red}{91\%} & 7\% & \textcolor{red}{92\%} & 5\% \\
        \hline
        \end{tabular}
        Falling is a sub-category of failure, and the failure rate = 1 - success rate
        \caption{The test results of many experiments (from 100 test episodes for each terrain).}
        \label{tab:final}
        \end{adjustwidth}
    \end{table}
    
    As previously discussed, this work uses a modified version of the \textit{OpenAI's BipedalWalker challenge}, meaning there are no available baselines and thus a baseline had to be developed. All experiments are conducted using the terrain mentioned in the section \ref{sec:terrain}. We use the original reward function for the developing \textbf{a blinded baseline policy}, which results in a behaviour where the agent attempts to travel through the terrain as fast as possible. Alternatively, as shown in Table \ref{tab:final}, with DRS the model has performed best across both experiment types: flat and rough terrain. Compared to the baseline policy, the DRS policy has about 90\% improvement overall. In rough terrain, there is a 6\% to 21\% improvement to a well-tuned static reward scheme. In flat terrain, DRS policy exhibits 2\% lower success rate than that of experiment 3, but is more than 10\% higher than the success rates of the experiment 1 and 4. Experiment 2 has a failed due to the target velocity not being achievable. From 100 testing episodes, with DRS, the agent performs equally as well with and without the terrain perturbation and has scored with 91\% and 92\% success rates, respectively. Therefore, we can conclude that the robot has learned a sufficiently generalisable locomotion model in this environment.
    
\begin{figure}[h]
        \centering
        \begin{center}
    	    \subfloat[The robot galloping on the rough terrain]{\includegraphics[width=0.8\textwidth]{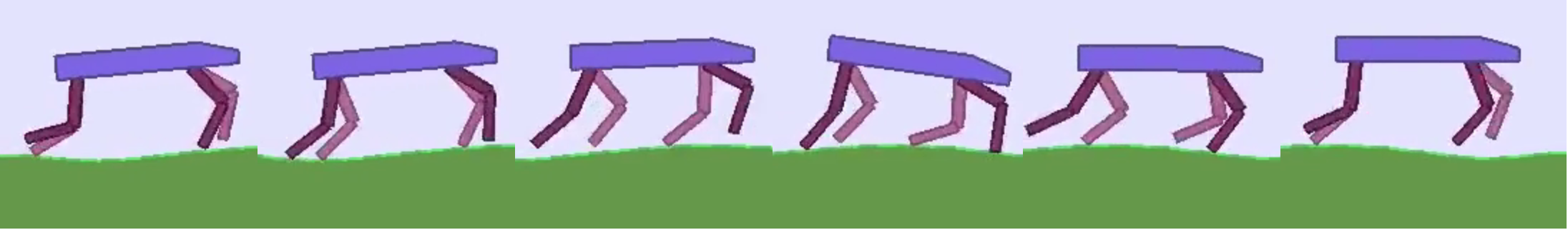}\label{fig:DRS_on_flat}}
    	\end{center} \hspace{0.2cm}
        \begin{center}
    	    \subfloat[The robot jumps over a gap]{\includegraphics[width=0.8\textwidth]{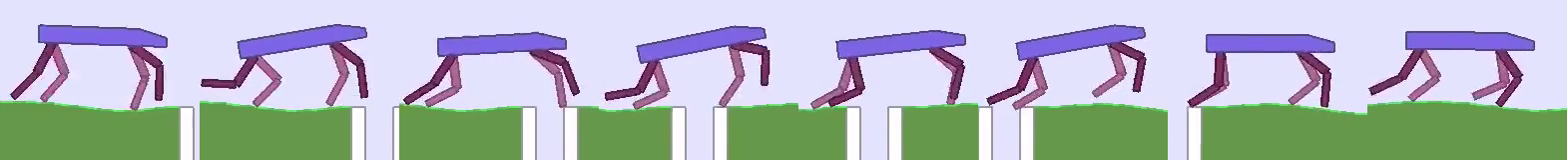}\label{fig:DRS_at_pit}}
    	\end{center} \hspace{0.2cm} 
        \begin{center}
    	    \subfloat[The robot jumps over a stump]{\includegraphics[width=0.8\textwidth]{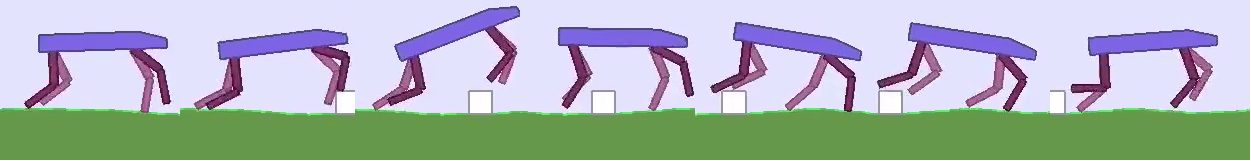}\label{fig:DRS_at_stump}}
    	\end{center} 
    	
        \caption{Successful robot locomotion trained using DRS on rough terrain.}
        \label{fig:DRS_gait}
    \end{figure}

    {\bf Ability of overcoming obstacles.} Without vision in challenging terrains, the state-of-the-art approaches must to reduce the travel velocity \cite{ETH} or getting ready to recovery from a fall \cite{yang2020}. With the aid of vision, a well-learned agent can overcome obstacles actively with ease (Figure \ref{fig:DRS_gait} and the attached video).
    
    
    {\bf The gait style.} As expected, using DRS, the robot displays all the desired properties presented in Section \ref{section:desired_gait_property}. The robot performs a smooth galloping gait with acceptable ability of body stabilization, as it can be seen in Figure \ref{fig:DRS_on_flat}. When encountering a gap, the robot actively anticipates the presence of the gap and steps over it, as shown in Figure \ref{fig:DRS_at_pit}. When encountering a stump, the robot actively jumps over it without first colliding against it or tipping over, as shown in Figure \ref{fig:DRS_at_stump}, where it can be seen that the robot jumps over a large stump easily. A video of a sample run is included in the supplementary materials.
    
    %
    %
    \begin{figure}[h]
        \centering
        \begin{center}
    	    \subfloat[The crawling gait]{\includegraphics[width=0.8\textwidth]{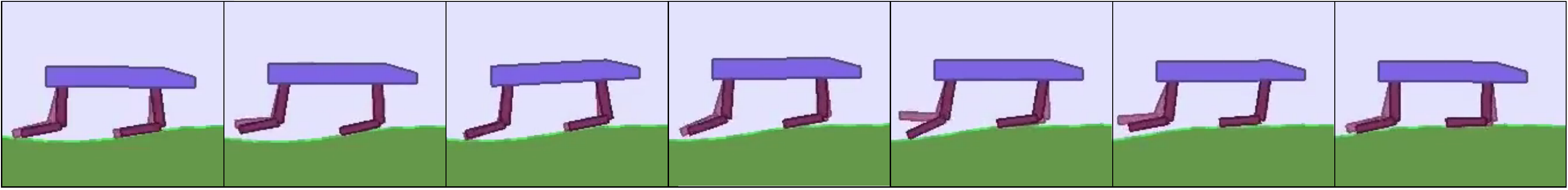}\label{fig:crawing agent}}
    	\end{center} \hspace{0.2cm} 
        \begin{center}
    	    \subfloat[The robot is being tripped by a small stump]{\includegraphics[width=0.8\textwidth]{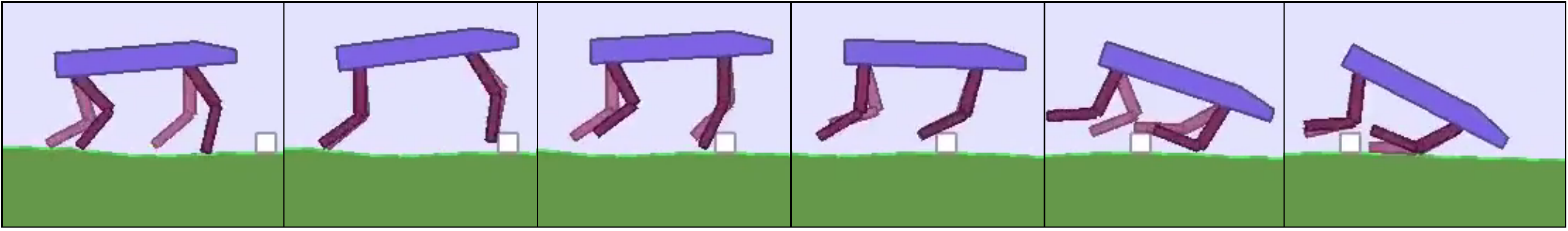}\label{fig:tripped by stump}}
    	\end{center} \hspace{0.2cm} 
        \begin{center}
    	    \subfloat[The robot is being stopped by a large stump but never felt down. The episode ends until reaching the maximum episode length.]{\includegraphics[width=0.8\textwidth]{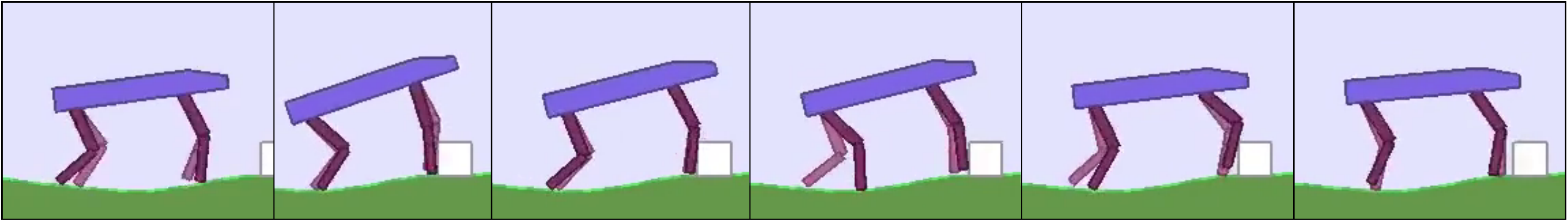}\label{fig:stuck at stump}}
    	\end{center} \hspace{0.2cm}
        \begin{center}
    	    \subfloat[The robot is failing at a gap.]{\includegraphics[width=0.8\textwidth]{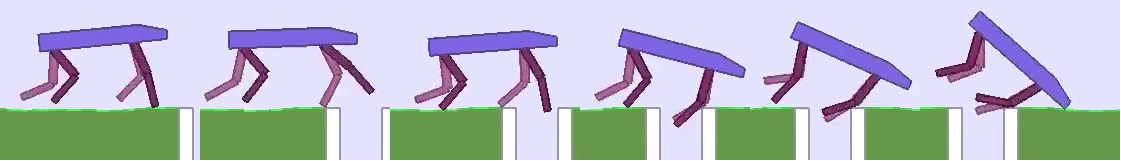}\label{fig:stuck at gap}}
    	\end{center} 
    	
        \caption{The failure gaits learned without DRS}
        \label{fig:bad_gaits}
    \end{figure}
    \clearpage
    \begin{figure}
        \centering
        \includegraphics[width=0.7\linewidth]{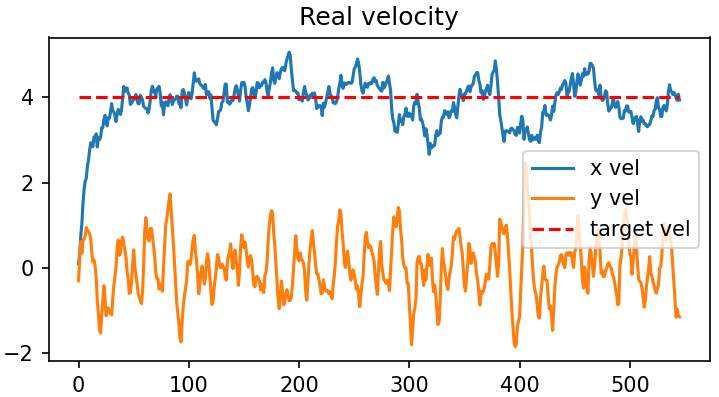}
        \caption{A sample of the velocity curve}
        \label{fig:DRS_vel_curve}
    \end{figure}
    
    {\bf The velocity smoothness.} As shown in Figure \ref{fig:DRS_vel_curve}, the horizontal velocity is very stable at 4~m/s. Even over obstacles, the deviation is still small. For the experiments, we only provide reference commands for the horizontal velocity. The vertical velocity is kept free for allowing better jumping behaviours. Comparing between the experiments 4 and 5, where a reference vertical velocity of 0 m/s is added, the gait becomes unnatural despite of the higher success rate, which is an exploitation of the simulation environment (see attached video). The constrained vertical target velocity may also result in failing gaits, see Figure \ref{fig:bad_gaits}. Especially in Figure \ref{fig:crawing agent}, at a low target velocity ($[2,0]^T$~m/s), the agent tends to crawl which is a sign of the policy being trapped into some local maxima. 
    
\begin{figure}[h]
        \centering
        \begin{subfigure}[T]{\textwidth}
            \centering
            \includegraphics[width=0.9\textwidth]{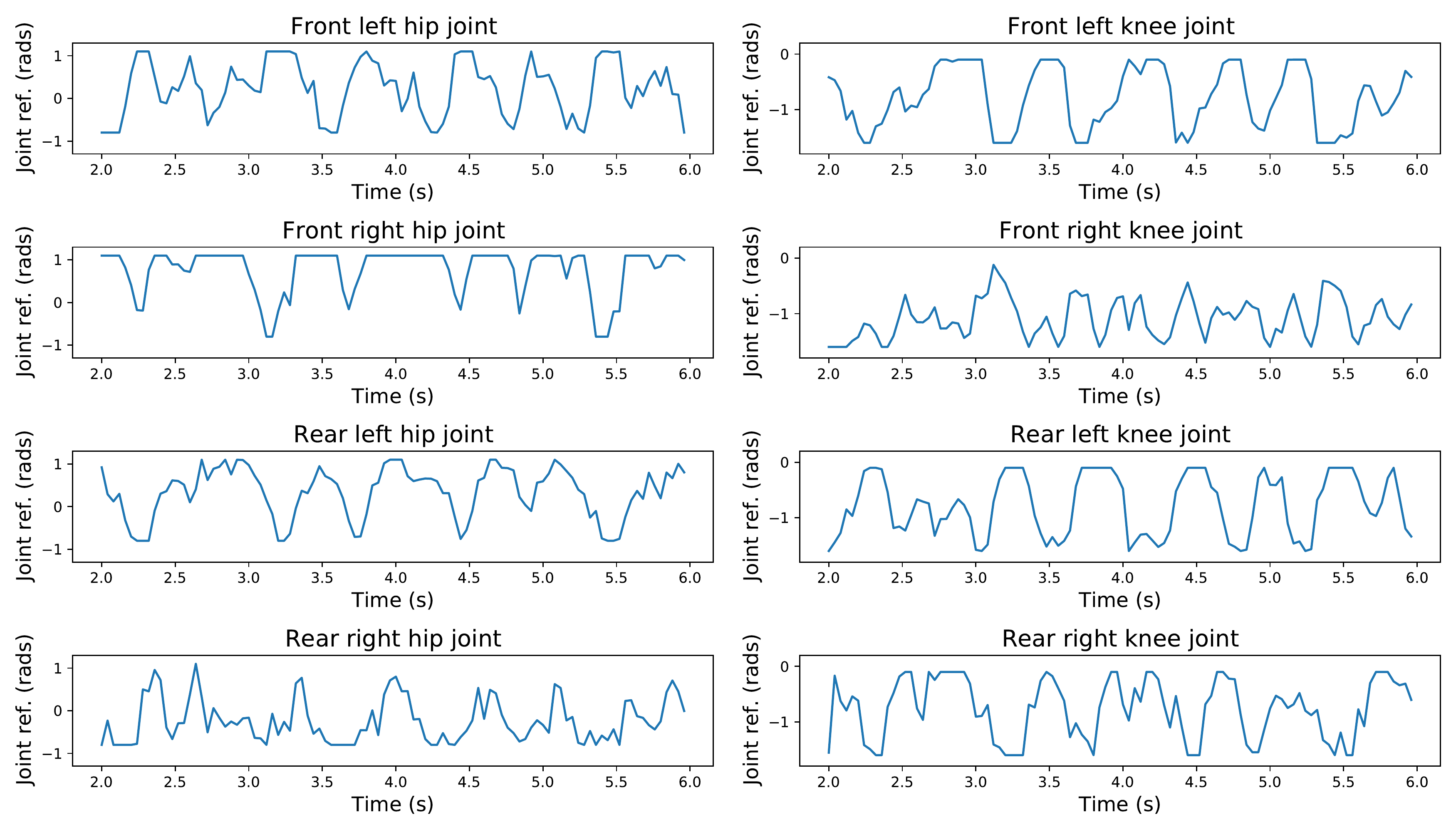}
            \caption{DRS policy 25Hz joint commands. \vspace{1em}}
            \label{fig:novel NN traj}
        \end{subfigure}
        \begin{subfigure}[T]{\textwidth}
            \centering \rule{0pt}{1cm}
            \includegraphics[width=0.9\textwidth]{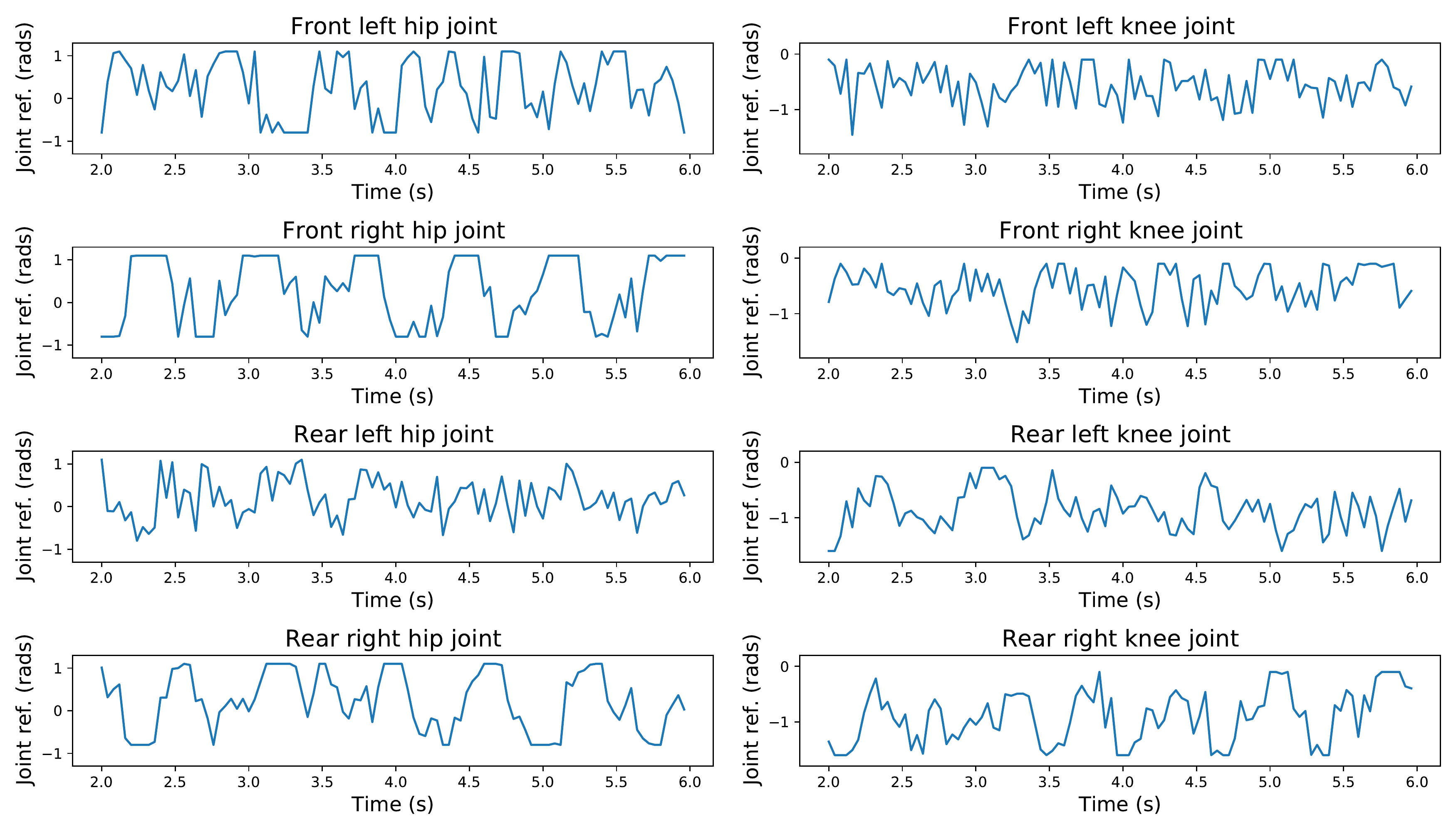}
            \caption{Baseline policy 25Hz joint commands.}
            \label{fig:shakyJointRefs} 
        \end{subfigure}
        
        
        \caption{Joint position commands of the policy trained with DRS and the baseline policy.}
        \label{fig:trajectory_quality}
    \end{figure}

    {\bf The trajectory quality.} With DRS, the policy generates less noisy and more periodic trajectories as shown in Figure \ref{fig:novel NN traj}. Given a smoother joint trajectory, the torques are more efficient, see the comparison in Figure \ref{fig:trajectory_quality}. 
    
    The major limitation of this project is the dimensionality. It is considerably easier to learn a locomotion model in 2-D than in 3-D. Because the dynamics of the robot becomes much more complicated in 3-D, and so does the terrain complexity. Besides, there are more possible gaits to learn, such as the orientation of the body, the diagonal$\slash$sideway gait, steering, foothold planning and more. Lastly, the approach proposed in this work requires a fine-tuned reward function and PD gains.

\section{Conclusion and Future Work}

    We present a vision-guided quadrupedal locomotion model formulated in a pure neural network architecture. The proposed model demonstrates its robust capability of \textit{active} obstacle avoidance by scoring at over 90\% success rate in all challenging terrains. Furthermore, the proposed Deep~RL heuristic Dynamic Reward Strategy (DRS) effectively learns a versatile gait using PPO on a customized quadrupedal robot in a modified OpenAI Gym simulation environment. This work can be served as a proof-of-concept to a vision-guided locomotion model to perform active terrain adaptation in 3-D. Although some privileged observations may not be available in the real-world, a synthesised observation information can be predicted as a latent vector by a combination of some sensory feedback \cite{kumar2021rma}. Finally, our DRS methodology could further boost the learning process under a more complex dynamics environment.
    

\clearpage

\appendix

\section{Hyperparameters for training the policy network}
\label{apx:hyperparm}
\begin{table}[h]
\begin{tabular}{llll}
    \textbf{MLP shape}: [256, 256]                  & \textbf{kl coefficient}: 1.0 \\
    \textbf{MLP activation function}: $tanh()$      & \textbf{kl target}: 0.01 \\
    \textbf{learning rate}: 0.00005                 & \textbf{value function clip parameter}: 10.0 \\
    \textbf{lambda}: 0.95                           & \textbf{stochastic gradient descent (ascent) minibatch size}: 20 \\
    \textbf{gamma}: 0.99                            & \textbf{stochastic gradient descent (ascent) iteration}: 20 \\
    \textbf{value function loss coefficient}: 0.62  & \textbf{train batch size}: 4000 \\
    \textbf{entropy coefficient}: 0.00045           & \textbf{sample fragment length}: 200 \\
    \textbf{clip parameter}: 0.3                    \\
\end{tabular}
\end{table}
%
%
\section{Parameters of the DRS}
\label{apx:parms}
\begin{table}[h]
\centering
\begin{tabular}{cccccccccc}
\hline\hline 
\addlinespace
{\bf Parameter} & $\mathrm{P}_{falling}$ & $v^{*}_{x}$ & $c_1$ & $c_2$ & $c_3$ & $c_4$ & $c_5$ & $c_6$ & $c_7$ \\
\addlinespace
\hline
\addlinespace
{\bf Value}     & -300 & 4.0 & 1.0 & 1.0 & 0.5 & 0.00096 & 0.00024 & 0.0024 & 0.0012 \\
\addlinespace
\bottomrule
\end{tabular}
\caption{Table of parameters for the novel reward function}\label{tab:rewardFuncParams}
\end{table}
\section{Proportional-Derivative parameters for the joint-level PD controller}
\begin{table}[h]
\centering
\begin{tabular}{ccccccccc}
\hline\hline 
 & \multicolumn{2}{c}{\textbf{Leg 0}} & \multicolumn{2}{c}{\textbf{Leg 1}} & \multicolumn{2}{c}{\textbf{Leg 2}} & \multicolumn{2}{c}{\textbf{Leg 3}} \\
\cmidrule{2-9}
 & Hip & Knee & Hip & Knee & Hip & Knee & Hip & Knee \\
\hline
\addlinespace
$K_p$~(Nm/rad)  & 100.0 & 100.0 & 100.0 & 100.0 & 100.0 & 100.0 & 100.0 & 100.0 \\
\addlinespace
$K_d$~(Nms/rad) & 2.0 & 1.0 & 2.0 & 1.0 & 2.0 & 1.0 & 2.0 & 1.0 \\
\addlinespace
\bottomrule
\end{tabular}
\caption{Table of the Proportional-Derivative parameters for the joint-level PD controller}\label{tab:PDParams}
\end{table}

\clearpage
\bibliography{main}

\end{document}